\documentclass[journal]{IEEEtran}

\usepackage{times}
\usepackage{epsfig}
\usepackage{graphicx}
\usepackage{amsmath}
\usepackage{amssymb}

\usepackage{float}
\usepackage{subfig}
\usepackage[flushleft]{threeparttable}
\usepackage{adjustbox}
\usepackage{tablefootnote}
\usepackage[font={normalsize}]{caption}
\usepackage{amsmath}
\usepackage{xcolor} 
\usepackage{dblfloatfix} 
\usepackage{booktabs,array}
\usepackage[detect-none]{siunitx}
\usepackage{multicol}
\usepackage[usestackEOL]{stackengine}
\usepackage{algorithmic}
\usepackage{algorithm}
\usepackage{hyperref}
\usepackage{wrapfig}
\usepackage{amssymb}
\usepackage{tabularx}
\usepackage{multirow}
\usepackage{makecell}
\usepackage{slashbox}

\newcolumntype{M}{>{$}c<{$}}
\newcolumntype{Z}{>{\centering\arraybackslash}X}
\newcolumntype{C}{>{\centering\arraybackslash}p}
\newcolumntype{Y}{>{\centering\arraybackslash}X}
\newcolumntype{L}{>{\raggedright\arraybackslash}p}

\ifCLASSINFOpdf
\else
\fi

\begin{document}

\title{Fingerprint Template Invertibility: Minutiae vs. Deep Templates}


\author{Kanishka~P.~Wijewardena,~Steven~A.~Grosz,~Kai~Cao~and~Anil~K.~Jain,~\IEEEmembership{Life~Fellow,~IEEE}}

\maketitle

\thispagestyle{empty}

\begin{abstract}
    \textbf{
     Much of the success of fingerprint recognition is attributed to minutiae-based fingerprint representation. It was believed that minutiae templates could not be inverted to obtain a high fidelity fingerprint image, but this assumption has been shown to be false. The success of deep learning has resulted in alternative fingerprint representations (embeddings), in the hope that they might offer better recognition accuracy as well as non-invertibility of deep network-based templates. We evaluate whether deep fingerprint templates suffer from the same reconstruction attacks as the minutiae templates. We show that while a deep template can be inverted to produce a fingerprint image that could be matched to its source image, deep templates are more resistant to reconstruction attacks than minutiae templates. In particular, reconstructed fingerprint images from minutiae templates yield a TAR of about 100.0\% (98.3\%) @ FAR of 0.01\% for type-I (type-II) attacks using a state-of-the-art commercial fingerprint matcher, when tested on NIST SD4. The corresponding attack performance for reconstructed fingerprint images from deep templates using the same commercial matcher yields a TAR of less than 1\% for both type-I and type-II attacks; however, when the reconstructed images are matched using the same deep network, they achieve a TAR of 85.95\% (68.10\%) for type-I (type-II) attacks. Furthermore, what is missing from previous fingerprint template inversion studies is an evaluation of the black-box attack performance, which we perform using 3 different state-of-the-art fingerprint matchers. We conclude that fingerprint images generated by inverting minutiae templates are highly susceptible to both white-box and black-box attack evaluations, while fingerprint images generated by deep templates are resistant to black-box evaluations and comparatively less susceptible to white-box evaluations.}\\
      
\end{abstract}

\section{Introduction}
    \IEEEPARstart{C}{ontinued}  advances in sensing, processing and storage technologies as well as fingerprint matching algorithms have contributed to fingerprint recognition's widespread adoption for reliable and secure person identification. Typically, fingerprint recognition algorithms rely on three classes of features extracted from a fingerprint image, shown in figure 1~\cite{handbook}:
    
    \begin{enumerate}
        \item Level 1: Global features such as pattern type, ridge orientation and frequency, and singular points (core and delta).
        \item Level 2: Local features, commonly referred to as minutiae, such as ridge endings and ridge bifurcations.
        \item Level 3: Fine-scale local features such as ridge contours, sweat pores, and incipient ridges.
    \end{enumerate}{}
    
    \begin{figure}
    \centering
    \includegraphics[width=\linewidth]{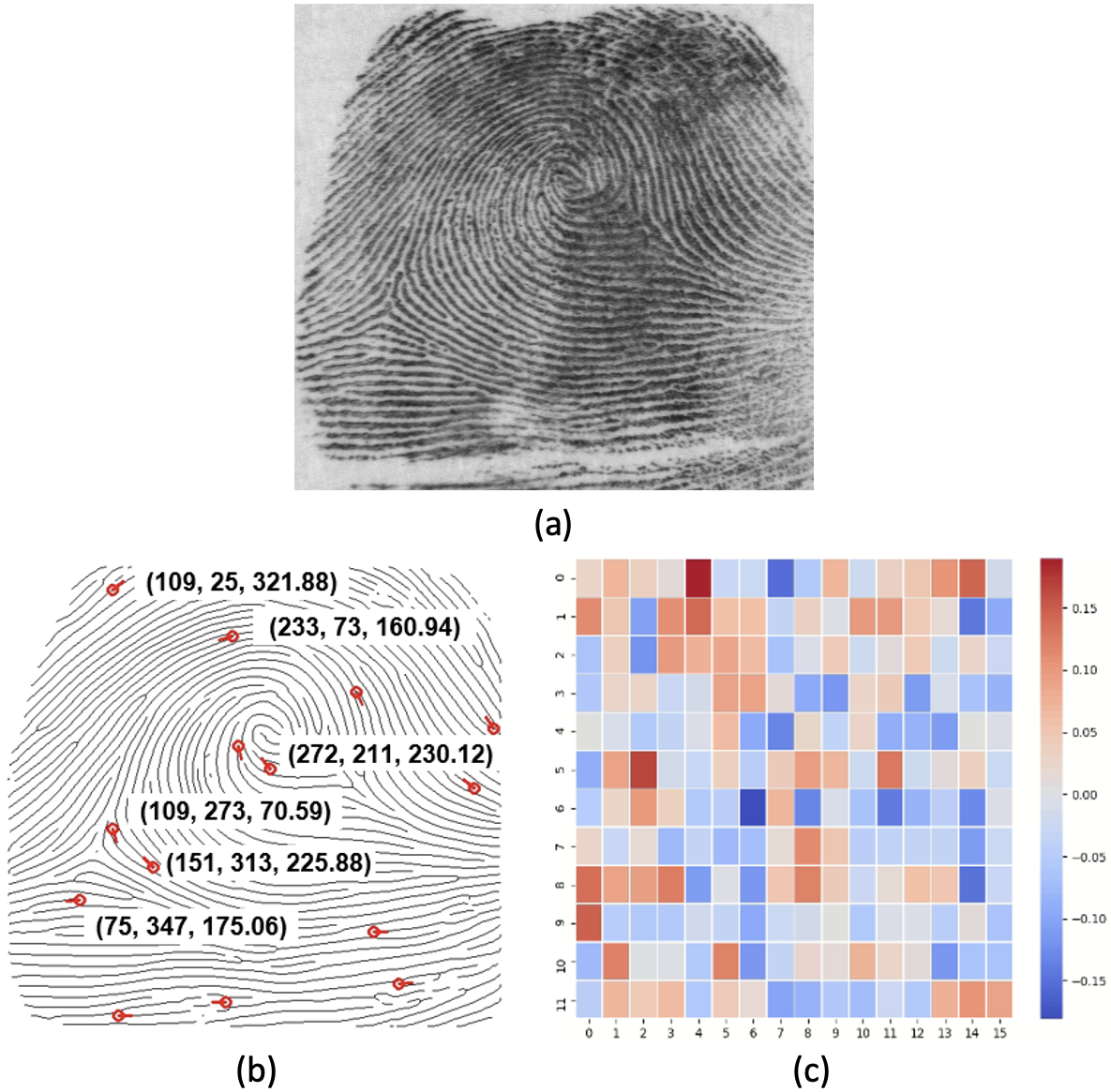} \\
    \caption{A sample fingerprint image from the NIST SD4 database (a); Corresponding skeleton image with some example minutiae points extracted by Verifinger v12.3 SDK annotated in ($x$, $y$, $\theta$) format (b); 192-dimensional deep template extracted by DeepPrint~\cite{engelsma2019learning} represented as a $12\times16$ heatmap (c).}
    \label{fig:fingerprintsample}
    \vspace{-1.5em}
    \end{figure}
    
    Minutiae points are considered to be the most discriminative characteristic of fingerprints and are the most widely used features in virtually all state-of-the-art (SOTA) fingerprint matching algorithms as well as by fingerprint examiners~\cite{fvc-ongoing}. At a minimum, each minutia point has an associated (x, y) spatial location within the image as well as an orientation corresponding to the direction of the ridge-line(s) associated with it. However, additional information such as minutiae type (e.g., ridge ending, bifurcation, etc.), quality, local descriptors, etc., may also be included and stored within a minutiae-template. Thus, to promote compatibility between minutiae-based fingerprint recognition vendors, the National Institute of Standard and Technology developed the international standard ISO/IEC 19794-2~\cite{ISO} to standardize commercial minutiae-based representations. In ISO/IEC 19794-2 formats, the origin of the coordinate system of the minutiae map is at the upper left corner of the original image with $x$ increasing to the right and $y$ increasing downward~\cite{Yoshida2010ASO}. ISO/IEC 19794-2:2005 contains three data formats for interchange and storage of fingerprint minutiae data: a record-based format, a normal format and compact formats, with optional extended data formats for including additional data such as ridge counts and core and delta locations. Even though the accuracy of fingerprint representations can be substantially increased by incorporating additional features, e.g., level III features such as dimensional attributes of the ridges, incipient ridges, breaks, creases and scars, such representations would involve using high quality sensing equipment, which in turn are computationally memory intensive for storage and computationally expensive when employed for matching algorithms~\cite{handbook}.
    
    Despite the success of minutiae-based representations for fingerprint recognition, there are alternative fingerprint representations that have been proposed to overcome some of the limitations inherent to minutiae-based matching, such as the varying length of minutiae-templates, latency in matching minutiae templates, storage requirements, securely protecting a minutiae template, etc. One such alternative is that of deep neural network-based representations (embeddings)~\cite{engelsma2019learning}~\cite{LiuFingerNet}, which are more compact, orders of magnitude faster during matching, and easier to encrypt due to their fixed-length nature and simple mathematical operations involved in matching two embeddings. Deep network embeddings are not just an alternative to traditional biometric feature-based representation, but can also complement traditional biometric templates such as in a multi-modal biometric template protection scheme~\cite{singh2021}. However, one important consideration which has yet to be examined for deep network-based fingerprint templates is the degree of invertibility of the embeddings; which, regardless of the fingerprint representation used, is a crucial factor threatening the security of fingerprint recognition systems. 
    
    Until about 20 years back, it was believed that a compromised minutiae feature-based template could not be inverted to successfully recover the source fingerprint image\footnote{In this work, the terms template reconstruction and template inversion are used interchangeably.}. Hence, reconstruction attacks on minutiae templates were not of a concern. However, a number of studies have now demonstrated that it is indeed possible to reconstruct a fingerprint image from a given minutiae set with sufficient fidelity to successfully match it with the fingerprint image from which the minutiae were extracted~\cite{cappelli2007fingerprint, ross2007template, feng2010fingerprint, li2012improved, cao2014learning, kim2019, bouzaglo2022synthesis}. The success of reconstruction techniques require that we secure fingerprint templates with strong encryption methods, develop strategies for matching fingerprints in the encrypted domain, or seek inherently irreversible representations. Toward that end, this paper aims to investigate whether deep-network-based fingerprint representations are invertible.
    
    Drawing from recent studies on template inversion attacks from deep templates in other biometric modalities, such as those by Mai et al.~\cite{mai2018reconstruction}, Ross et al.~\cite{rossvein2020} and Ahmad et al.~\cite{Ahmad2007}, it seems possible that the susceptibility of template inversion attacks on fingerprint embeddings may be of concern. However, the question of invertibility of fingerprint embeddings has yet to be studied and compared to the degree of success of inversion attacks on minutiae-based representations. In order to provide a fair comparison between minutiae inversion attacks and inversion attacks on deep network fingerprint embeddings, we first design our own convolutional neural network to invert minutiae-based representations and later modify the architecture to invert deep template embeddings instead. The motivation to design our own minutiae inversion network, rather than rely on past results reported in the literature, is three fold: (i) past studies either use out-dated reconstruction methods which leave room for improvement in terms of attack success, (ii) incorporate auxiliary information outside of just the spatial location and orientation ($x$, $y$, and $\theta$) of minutiae points, or (iii) use out-dated fingerprint matching systems to assess the attack performance - which make a comparison difficult and does not accurately reflect the threat of inversion attacks today. This work also investigates the success of inverting a fingerprint template (minutiae-based or deep network-based) produced by one fingerprint template extractor in attacking another, unrelated fingerprint matching system (e.g., a black-box attack). Specific contributions of this work include:
    
    \begin{enumerate}
        \item An evaluation on the invertibility of fingerprint templates produced by deep networks. To the best of our knowledge, this is the first work on image reconstruction from deep templates applied to fingerprint recognition.
        \item An improved minutiae template inversion algorithm, which seeks to reconstruct fingerprint images from just the spatial location and orientation (x, y, and $\theta$) of the minutiae points.  
        \item Analysis of template inversion attack performance across matchers, including those not used during training of our inversion algorithm. According to our knowledge, this is the first work that evaluates the accuracy of fingerprint image reconstruction from multiple minutiae-based fingerprint template generation schemes by matching the images generated by one minutiae-based template scheme with their source images using another fingerprint matcher. For this work, we define the template reconstruction attack performed by a matcher upon its own template generation SDK as a white-box attack, and we also define the template reconstruction attack performed by a matcher upon a different template generation SDK as a black-box attack. 
    \end{enumerate}
    
    \begin{figure}
    \begin{center}
    \includegraphics[width=\linewidth]{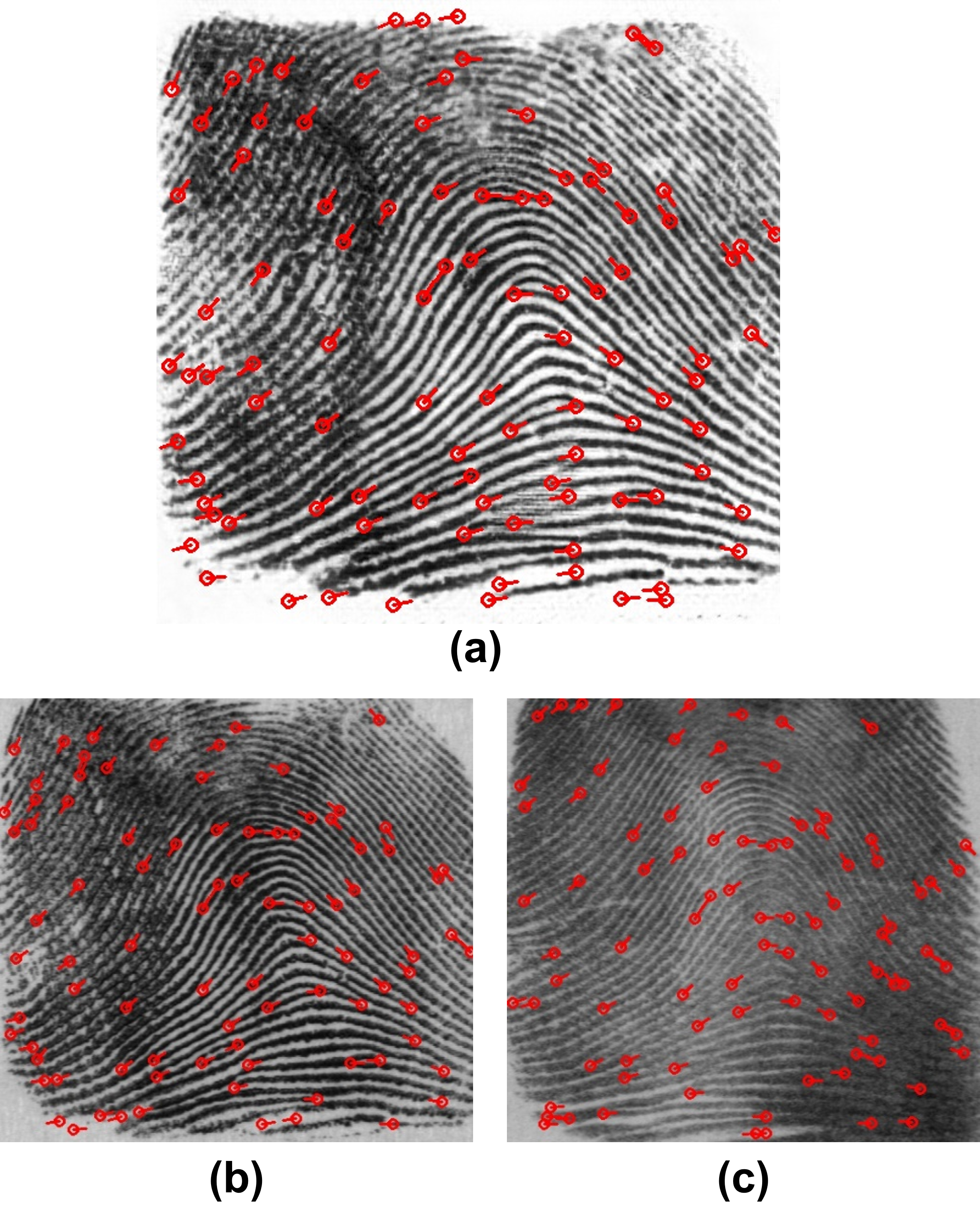}
    \caption{Illustration of type-I and type-II attacks, featuring: A reconstructed fingerprint image from a minutiae template (a); The source image from NIST SD4 from which the minutiae template was extracted (b); A different impression of the source fingerprint from NIST SD4 (c)}
    \label{fig:attack_types}
    \end{center}
    \vspace{-1.5em}
    \end{figure}

\section{Related Work}

\begin{table*}[t]
\small
\renewcommand{\arraystretch}{1.3}
\caption{Comparison of Previous Fingerprint Image Reconstruction Approaches from Minutiae Templates}
\label{prior_work}
\centering
\begin{tabular}{||m{0.08\linewidth}|m{0.16\linewidth}|m{0.38\linewidth}|m{0.27\linewidth}||}
\noalign{\hrule height 1.5pt}
\textbf{Algorithm} & \textbf{Method} & \textbf{Performance Evaluation} & \textbf{Comments} \\
\noalign{\hrule height 1.0pt}
Cappelli et al.~\cite{cappelli2007fingerprint} & Modified zero-pole model with Gabor filtering & Type-I attack: 81.49\% TAR at 0\% FAR on FVC2002 DB1\_A & Suffers from spurious minutiae in the reconstructed image\\
\hline
Ross et al.~\cite{ross2007template} & Using minutiae triplets for orientation field estimation and using streamlines and line integral convolution for ridge pattern reconstruction & Type-I attack has a 23\% identification rate at the rank-1 level on NIST SD4 & Only partial fingerprint images could be reconstructed\\
\hline
Feng \& Jain~\cite{feng2010fingerprint} & Orientation estimated using nearest minutiae and AM-FM model & Type-I (type-II) attack: 94.13\% (45.89\%) TAR at 0.1\% FAR on FVC2002 DB1\_A. Type-I (type-II) rank-1 identification rate: 99.70\% (65.75\%)  on NIST SD4 & Spurious minutiae present in high curvature regions, requires a better model for the continuous phase\\
\hline
Li \& Kot~\cite{li2012improved} & Orientation estimated using nearest minutiae and AM-FM model & Type-II attack: 86.48\% TAR at 0.01\% FAR on FVC2002 DB1\_A; 86.96\% TAR at 0.01\% FAR on FVC2002 DB2\_A & Ridge pattern may differ in the continuous phase compared to the original fingerprint image\\
\hline
Cao \& Jain~\cite{cao2014learning} & Orientation patch and continuous phase patch dictionaries & Type-I (type-II) attack: 100\% (85.23\%) TAR at 0.01\% FAR on FVC2002 DB1\_A. Type-I (type-II) attack: 100\% (90.29\%) TAR at 0.01\% FAR on FVC2002 DB2\_A & Requires learning orientation and ridge pattern for reconstruction. Fingerprints are ideal and without ``noise", thus cannot fool a human expert.\\
\hline
Bouzaglo \& Keller~\cite{bouzaglo2022synthesis} & Used StyleGAN2, with a minutiae-to-vector encoder & Type-I (type-II) attack: 99.26\% (86.32\%) TAR at 0.01\% FAR on NIST SD4 & Incorporates information regarding minutiae location, orientation, and class for fingerprint reconstruction. Utilized 54,000 training images from NIST SD14\\
\hline
\textbf{Proposed Work} & Comparison of inversion of minutiae-based templates vs. fixed-length deep fingerprint representations & Minutiae inversion: TAR$>88\%$ at FAR of 0.01\% for type-II attacks on NIST SD4 using COTS Matchers. DeepPrint inversion: TAR of 68.10\% at FAR of 0.01\% for type-II attacks on NIST SD4 using DeepPrint matcher & DeepPrint inversion did not result in successful matching via COTS matchers, with TAR$<1\%$ at FAR of 0.01\% for inversion on NIST SD4 \\
\noalign{\hrule height 1.5pt}
\end{tabular}
\end{table*}

\subsection{Fingerprint Minutiae Template Inversion}

    Initial attempts at image reconstruction from fingerprint templates originate from Hill in~\cite{hill2001risk}. The reconstruction method proposed by Hill incorporates a zero-pole model to reconstruct the fingerprint orientation field from core and delta singular points. Cappelli et al.~\cite{cappelli2007fingerprint} proposed a variant of the zero-pole model to incorporate the minutiae direction information. This produced a very coarse ridge pattern connecting various minutiae that was able to produce a significant enough similarity match score to many of the source fingerprints from which singular points and minutiae were extracted. However, this approach was not applicable to the case when singular points are not present in the fingerprint template. This problem was addressed in Ross et al.~\cite{ross2007template}, where triangles of minutiae triplets were used to reconstruct orientation fields without using singular points. 
    
    In another approach by Feng and Jain~\cite{feng2010fingerprint}, the authors predicted an orientation value within each local image block using the nearest minutia in the eight surrounding sectors, based on the AM-FM model utilized for fingerprint representation. Li \& Kot~\cite{li2012improved} also reconstructed fingerprint ridges utilizing the AM-FM model for fingerprint representation. This work proposed an approach that improved upon the accuracy of matching a reconstructed fingerprint image to a different impression of the same fingerprint, compared to Feng and Jain~\cite{feng2010fingerprint}. Cao and Jain~\cite{cao2014learning} used dictionaries of orientation patches and ridge structure to further improve the fingerprint reconstruction. 
    
    With the increased use of deep learning tools in Biometrics and related fields, it wasn't long before Generative Adversarial Networks were utilized for fingerprint image reconstruction from minutiae points. Kim et al.~\cite{kim2019} utilized a Conditional Generative Adversarial Network to reconstruct fingerprint images from their corresponding minutiae maps by first extracting the minutiae, and then generating an image of a minutiae map, which was then trained on a conditional GAN architecture to output a reconstructed fingerprint image. In addition, Bouzaglo and Keller~\cite{bouzaglo2022synthesis}, used the StyleGAN2 architecture combined with a minutiae-to-vector encoder, to synthesize and reconstruct fingerprint images from minutiae templates. The minutiae templates contained x, y, $\theta$ (minutiae direction) and T (minutiae class) information. The fingerprint minutiae templates were first extracted using a COTS matcher after which they were converted into minutiae maps that were then fed into a Minutiae-to-Vector encoder. The inclusion of minutiae class in the fingerprint template provided additional information that aids in the construction of the subsequent minutiae map. Thus, such minutiae maps are less likely to lose precision than when constructing minutiae maps from just the x, y and $\theta$ information. 
    
\subsection{Deep-Network Fingerprint Embeddings}
    Despite the success of minutiae-based fingerprint representations in achieving extremely high verification accuracy, many recent works have explored the use of deep network-based fingerprint embeddings as an alternative or auxiliary template to the classical minutiae representation~\cite{LiuFingerNet}. Deep fingerprint embeddings offer several advantages over the classical minutiae representation; mainly, improved computational and storage efficiency due to the potential for compact, fixed-length representations. These fixed-length representations do not require computationally expensive graph matching techniques; therefore, are significantly faster during matching and 1:N search. They have also been shown to have better discriminative power in poor quality fingerprints where minutiae extraction is unreliable~\cite{engelsma2019learning}.
    
    One recent deep network-based fingerprint representation proposed in the literature (and used in this study) is that of DeepPrint, a deep learning-based approach for fingerprint representation proposed by Engelsma et al.~\cite{engelsma2019learning}. DeepPrint is able to produce highly discriminative fingerprint embeddings by incorporating the traditional minutiae-based fingerprint domain knowledge in the learning of a fixed length representation of just 200 bytes. When a DeepPrint fingerprint is enrolled into a system, it is first aligned via a Localization Network which has been trained end-to-end with a base network and feature extraction networks. Then, the aligned fingerprint proceeds to the base network which is followed by two branches; the first branch extracts the texture-based representation; and the second branch extracts the minutiae-based representation, guided by a side-task of minutiae detection (via a minutiae map which does not have to be extracted during testing). The texture-based representation and the minutiae-based representation are concatenated into a 192-dimensional representation of 768 bytes (192 features and 4 bytes per float). The 768-byte template is then compressed into the final 200-byte fixed-length representation. 
    
\subsection{Inversion Attacks on Other Biometric Modalities}
    There has been some work done in the past to assess the success of reconstructing a source image from a deep-network biometric template representation. In Ross et al.~\cite{rossvein2020}, grayscale finger vein images were reconstructed from a template consisting of binary features. This was achieved using a convolutional neural network, which could then be used for cross-dataset reconstruction and biometric recognition purposes. In Ahmad et al.~\cite{Ahmad2007}, an inversion-based on a convolutional neural network architecture called RESIST (REconStructing IriSes from Templates) was applied to reconstruct iris images from 3 different feature extraction pipelines. The architecture performed well on all three types of templates, with deep templates proving harder to invert than templates generated using Gabor filtering. A limitation in this approach was that the inversion models were only trained on 64,980 iris samples from 356 different subjects.
    
    In the realm of using deep-network templates in face recognition, Zhmoginov and Sandler~\cite{zhmoginov2016inverting} apply both an iterative reconstruction approach and a feed-forward network approach to learn the reconstruction of face images by minimizing the difference between source and reconstructed templates. The authors were able to show that the reconstructed images resembled the faces of human subjects visually similar to the faces of the target image. In another study, Cole et al.~\cite{cole2017synthesizing} estimated landmarks and textures of face images from deep templates, and then combined the estimated features using a differentiable warping to yield reconstructed images. However, this work did not specifically address the issue of face reconstruction attacks on face templates. Finally, Mai et al.~\cite{mai2018reconstruction} employed neighborly de-convolutional neural networks toward the task of reconstructing deep network embeddings of face images and achieved a true acceptance rate (TAR) of 95.20\% (58.05\%) on LFW under type-I (type-II) \footnote{Type-I attack: matching a reconstructed image of a fingerprint/face against the same fingerprint impression/face image from which the template embedding was extracted. Type-II attack: matching a reconstructed fingerprint/face image against a different impression of the same finger/different face image of the same subject} attacks at a false acceptance rate (FAR) of 0.1\%.

\section{Proposed Template Inversion Approaches}
    Many previous methods have demonstrated success in inverting some variation of a minutiae-based fingerprint template; However, most incorporate additional information outside of just the spatial location and orientation of the minutiae points or utilize outdated methods (modern CNN-based template inversion methods can achieve a TAR$\>99\%$ @ FAR of 0.01\%) for either the reconstruction itself or outdated matchers to assess the performance of the inversion attacks. Therefore, we first develop our own minutiae-based fingerprint inversion network to establish a better baseline to compare with the invertibility of deep network-based fingerprint embeddings. We then slightly modify the architecture to invert deep network-based embeddings rather than minutiae.

\subsection{Minutiae template inversion}
    To invert minutiae templates, we first convert the list of x, y, and $\theta$ information to minutiae maps which are better suited for input to a convolutional neural network. To obtain minutiae maps for training, we use the CNN-based minutiae extractor proposed by Cao et al.~\cite{KaiMinutiae}, hereby referred to as MSU-LatentAFIS as it can easily be embedded into a fully differential learning framework and used as a supervision for guiding the template reconstruction network. When applying our reconstruction network to minutiae templates obtained from any other minutiae extraction SDKs (e.g., Verifinger v12.3 SDK or Innovatrics ANSI\&ISO v2.4.10 SDK) we simply convert the minutiae (x, y, and $\theta$) information to minutiae maps as input to our algorithm.
    
\begin{figure*}
    \begin{center}
    \includegraphics[width=\linewidth]{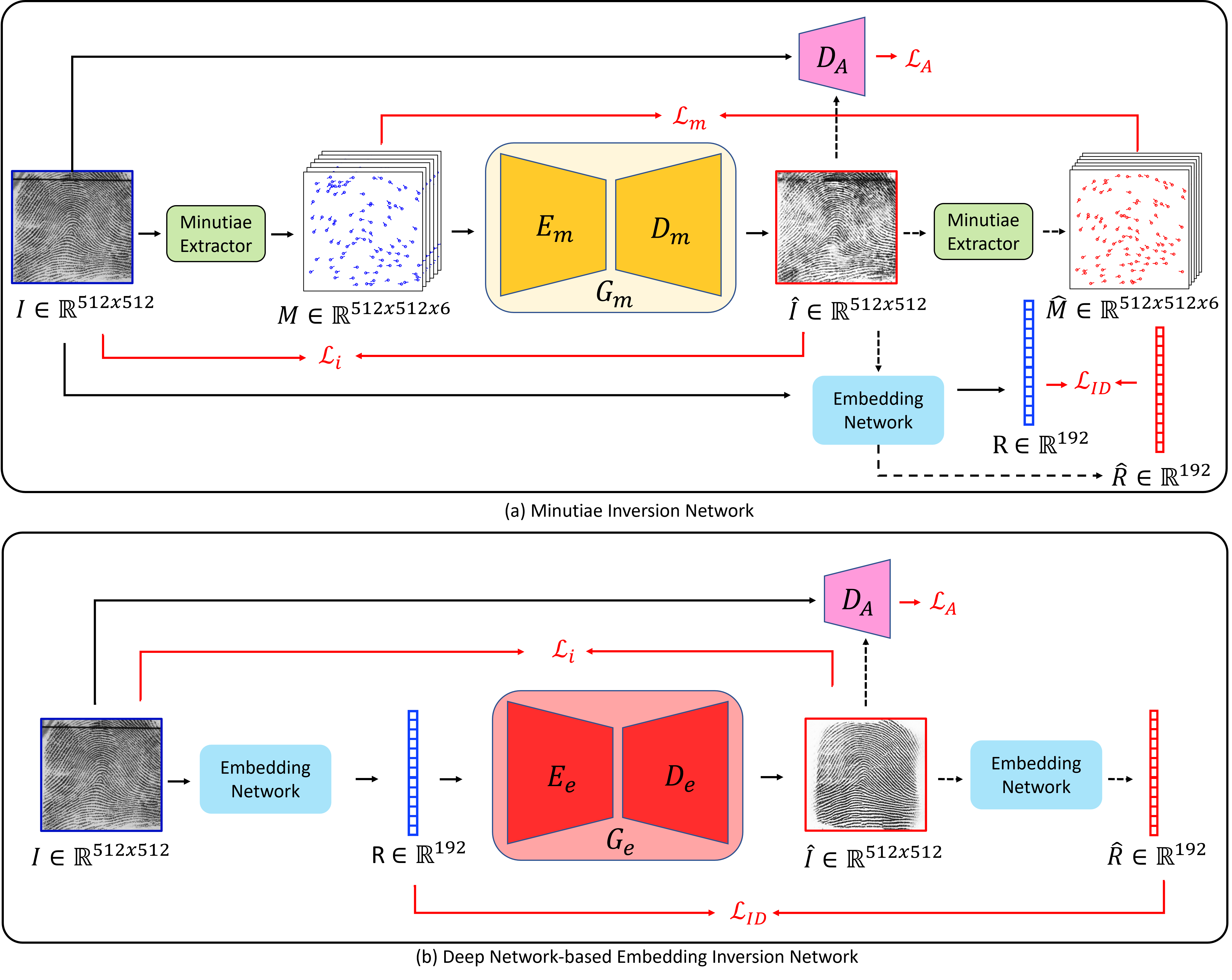}
    \caption{Overview of architecture used for minutiae template to image inversion (a), and deep network embedding to image inversion (b)}
    \label{fig:mmap_overview}
    \end{center}
    \vspace{-1.5em}
    \end{figure*}

    Our network for reconstructing fingerprint images from minutiae follows an encoder-decoder ($E_m$ and $D_m$) architecture $G_m(\cdot)$, in which input minutiae maps $M(x,y,\theta)$, are converted to grayscale fingerprint images. The encoder-decoder architecture utilizes ``Residual Blocks'' from the BigGAN architecture~\cite{BigGAN}\footnote{BigGAN uses ``Residual Blocks" in its Generator and Discriminator. For detailed information regarding these please refer to Brock et al.~\cite{BigGAN}}. Our overall training process involves multiple losses to both ensure realistic fingerprint reconstructions and preserve the identity in the reconstructed fingerprint images. A typical GAN loss is utilized to constrain the reconstructions to produce convincing fingerprint images using a discriminator, $D_A(\cdot)$, which penalizes non-realistic fingerprint images according to the classical GAN loss, $\mathcal{L}_{A}$, (Eq.~\ref{eq:gan_loss}). The losses utilized to preserve the identity of the generated fingerprints include a DeepPrint identity loss, $\mathcal{L}_{ID}$, (Eq.~\ref{eq:dp_loss}), a minutiae identity loss, $\mathcal{L}_{m}$, (Eq.~\ref{eq:mmap_loss}), and an image/pixel reconstruction loss, $\mathcal{L}_{i}$, (Eq.~\ref{eq:image_loss}). An additional $L2$ regularization loss, $\mathcal{L}_{reg}$, on the weights of the generator was also used. The individual losses are explained in more detail below. The overall loss used to update the parameters of $G_m(\cdot)$ is given in Eq.~\ref{eq:overall_loss_G} and the loss for updating $D_A(\cdot)$ is given in Eq.~\ref{eq:overall_loss_D}. The architectural details of $G_m(\cdot)$ and $D_A(\cdot)$ are given in Tables \ref{mmap_inversion_architecture} and \ref{mmap_discriminator_architecture}; respectively. A high-level overview of the approach is given in (a) of Figure~\ref{fig:mmap_overview}.

    \begin{enumerate}
        \item GAN loss ($\mathcal{L}_{A}$): Classical min-max GAN loss between the discriminator, $D_A(\cdot)$, trying to classify each source fingerprint image, $I$, as real and each image reconstructed from minutiae template inversion $\hat{I}=G_m(M)$ as fake. Meanwhile, $G_m(\cdot)$ is trying to fool $D_A(\cdot)$ into thinking its outputs come from the source image distribution.
            \begin{equation}
            \label{eq:gan_loss}
            \mathcal{L}_{A} =  \mathbb{E}_I~[logD_A(I)] + \mathbb{E}_M~[log(1-D_A(G_m(M)))]
            \end{equation}
        \item Orthogonal regularization loss ($\mathcal{L}_{reg}$): $L2$ distance between the Gram matrix of the orthogonally initialized weights of the model $W$ and the identity matrix $I$, multiplied by a scaling factor $\beta = 0.0001$. 
            \begin{equation}
            \label{eq:ortho_loss_dp}
            \mathcal{L}_{A} = \beta||W^TW\odot(1-I)||^2_F
            \end{equation}
        \item Minutiae Map loss ($\mathcal{L}_{m}$): $L1$ distance between ground truth minutiae maps, $M$, and minutiae maps extracted from corresponding reconstructed images, $\hat{M}$.
            \begin{equation}
            \label{eq:mmap_loss}
            \mathcal{L}_m = \sum_{i,j,k} |M(x_i,y_j,\theta_k)-\hat{M}(x_i,y_j,\theta_k)|
            \end{equation}
        \item DeepPrint loss ($\mathcal{L}_{ID}$): $L2$ distance between the DeepPrint embedding extracted from the source fingerprint image, $R$, and the DeepPrint embedding extracted from the reconstructed fingerprint image, $\hat{R}$. 
            \begin{equation}
            \label{eq:dp_loss}
            \mathcal{L}_{ID} = \frac{1}{2}\sum(R - \hat{R})^2 \\
            \end{equation}
        \item Image/pixel loss ($\mathcal{L}_{i}$): $L2$ loss between the source fingerprint image, $I$, and reconstructed fingerprint image, $\hat{I}$.
            \begin{equation}
            \label{eq:image_loss}
            \mathcal{L}_i = \frac{1}{2}\sum_{x,y} (I(x,y) - \hat{I}(x,y))^2
            \end{equation}
        \item Overall loss ($\mathcal{L}_{gen_m}$) for $G_{m}(\cdot)$: $\lambda_{1}=1$, $\lambda_{2}=2$, $\lambda_{3}=1$, and $\lambda_{4}=10$. 
            \begin{equation}
            \label{eq:overall_loss_G}
            \mathcal{L}_{gen_m} = \lambda_1\mathcal{L}_{A} + \lambda_2\mathcal{L}_{m} + \lambda_3\mathcal{L}_{ID} + \lambda_4\mathcal{L}_{i}
            \end{equation}
        \item Overall loss ($\mathcal{L}_{disc_m}$) for $D_A(\cdot)$:
            \begin{equation}
            \label{eq:overall_loss_D}
            \mathcal{L}_{disc_m} = \mathcal{L}_{A}
            \end{equation}
    \end{enumerate}

    \begin{table}
    \renewcommand{\arraystretch}{1.3}
    \caption{Minutiae map to image inversion encoder-decoder architecture}
    \label{mmap_inversion_architecture}
    \centering
     \begin{tabular}{>{\centering\arraybackslash}m{0.25\textwidth}||>{\centering\arraybackslash}m{0.15\textwidth}}
    \hline
     \textbf{Layer} & \textbf{Output Dimensions} \\
    \hline
    \hline
    Input & $512 \times 512 \times 6$ \\
    \hline
    Enc 0: ResBlock Down + ResBlock  & $256 \times 256 \times 48$ \\
    \hline
    Enc 1: ResBlock Down + ResBlock  & $128 \times 128 \times 96$ \\
    \hline
    Enc 2: ResBlock Down + ResBlock  & $64 \times 64 \times 192$ \\
    \hline
    Non-Local Block $(192 \times 192)$  & $64 \times 64 \times 192$ \\
    \hline
    Enc 3: ResBlock Down + ResBlock  & $32 \times 32 \times 384$ \\
    \Xhline{4\arrayrulewidth}
    Dec 0: ResBlock Up + ResBlock  & $64 \times 64 \times 384$ \\
    \hline
    Dec 1: ResBlock Up + ResBlock  & $128 \times 128 \times 192$ \\
    \hline
    Non-Local Block $(192 \times 192)$  & $128 \times 128 \times 192$ \\
    \hline
    Dec 2: ResBlock Up + ResBlock  & $256 \times 256 \times 96$ \\
    \hline
    Dec 3: ResBlock Up  & $512 \times 512 \times 48$ \\
    \hline
    Batch Normalization, ReLU  & $512 \times 512 \times 48$ \\
    \hline
    Convolution (channels=1, kernel=$3\times3$, stride=1) & $512 \times 512 \times 1$ \\
    \hline
    Tanh & $512 \times 512 \times 1$ \\
    \Xhline{2\arrayrulewidth}
    \end{tabular}
    \end{table}
    
    \begin{table}
    \renewcommand{\arraystretch}{1.3}
    \caption{Minutiae map to image inversion discriminator architecture}
    \label{mmap_discriminator_architecture}
    \centering
     \begin{tabular}{>{\centering\arraybackslash}m{0.25\textwidth}||>{\centering\arraybackslash}m{0.15\textwidth}}
    \hline
    \textbf{Layer} & \textbf{Output Dimensions} \\
    \hline
    \hline
    Input & $512 \times 512 \times 1$ \\
    \hline
    Disc 0: ResBlock Down & $256 \times 256 \times 48$ \\
    \hline
    Disc 1: ResBlock Down & $128 \times 128 \times 96$ \\
    \hline
    Disc 2: ResBlock Down & $64 \times 64 \times 192$ \\
    \hline
    Non-Local Block $(192 \times 192)$  & $64 \times 64 \times 192$ \\
    \hline
    Disc 3: ResBlock Down & $32 \times 32 \times 384$ \\
    \hline
    Disc 4: ResBlock Down & $16 \times 16 \times 768$ \\
    \hline
    ReLU  & $16 \times 16 \times 768$ \\
    \hline
    Global Summation Pooling  & $768$ \\
    \hline
    Linear  & $1$ \\
    \Xhline{2\arrayrulewidth}
    \end{tabular}
    \end{table}
    
    For training the minutiae-map to image inversion model, we used a collection of 447,988 rolled fingerprint impressions from an operational, forensic fingerprint dataset introduced by Yoon et al. in~\cite{MSP_longitudinal}, referred to as the Michigan State Police (MSP) Longitudinal fingerprint dataset. The source images were captured as $480\times512$ images at a resolution of 500 dpi, which were padded to a size of $512\times512$ pixels. An additional plain fingerprint dataset collection, comprising of 7,901 images from the MSU Presentation Attack Dataset (MSU-FPAD) was also used for training the minutiae map inversion model~\cite{chugh2018fingerprint}. These images were captured as $750\times800$ images at a resolution of 500 dpi, which were then segmented and padded to a size of $512\times512$ pixels. We extracted 6-channel minutiae maps for the entire training set for input to our network, which yielded an input minutiae map of $512\times512\times6$ dimensions.
    
    The models utilized for inverting from MSULatent-AFIS, Verifinger and Innovatrics templates were trained on a server containing two RTX 1080 TI GPUs, each containing 11 GB of memory. The generator learning rate was set at 0.0001 and the discriminator learning rate was set at 0.0001. The Generator was updated for three iterations for each iteration the Discriminator was updated. 

\subsection{Deep-Network Template Inversion}
    After inverting the minutiae maps to obtain the reconstructed fingerprints, a new inversion model was trained to reconstruct images from deep fingerprint templates (i.e., embeddings). As mentioned previously, we sought to invert deep template fingerprint embeddings, originally proposed in~\cite{engelsma2019learning}, for our inversion experiments.

    Our deep template inversion model follows an encoder-decoder architecture. Both the minutiae template inversion model and the deep template inversion model share the same decoder network, with slight changes to the encoder network given the different inputs into each network. Concretely, our network takes as input a 192-d vector and reconstructs the source fingerprint image at a resolution of $512\times512$.

    An overview of the DeepPrint to fingerprint image inversion algorithm is given in (b) of Figure~\ref{fig:mmap_overview}. The fingerprint reconstruction generator $G_{e}(\cdot)$ consists of an encoder-decoder ($E_e$ and $D_e$). Similar to the minutiae map inversion process, a discriminator, $D_{A}(\cdot)$, is used to provide the GAN loss, $\mathcal{L_A}$, in supervising realistic fingerprint reconstructions. In addition, an $L2$ DeepPrint identity loss, $\mathcal{L}_{ID}$, between the source image DeepPrint embedding, $R$, and the reconstructed image DeepPrint embedding, $\hat{R}$, and an $L1$ image/pixel reconstruction loss, $\mathcal{L}_{i}$, were incorporated during training. Minutiae-based identity loss $\mathcal{L}_{m}$ was not incorporated into this model as it did not improve the performance of the model against COTS matchers (our black-box evaluations), and significantly worsened the model's performance in the white-box setting. An additional $L2$ orthogonal regularization loss on the weights of the generator was used. The total list of losses used to train the DeepPrint inversion network is given below. 
    
    \begin{enumerate}
        \item GAN loss ($\mathcal{L}_{A}$): Classical min-max GAN loss between the discriminator, $D_A(\cdot)$, trying to classify each source fingerprint image, $I$, as real and each deep template inversion $\hat{I}=G_e(R)$ as fake. Meanwhile, $G_e(\cdot)$ is trying to fool $D_A(\cdot)$ into thinking its outputs come from the source image distribution.
            \begin{equation}
            \label{eq:gan_loss_dp}
            \mathcal{L}_{A} =  \mathbb{E}_I~[logD_{A}(I)] + \mathbb{E}_R~[log(1-D_{A}(G_{e}(R)))]
            \end{equation}
        \item Orthogonal regularization loss ($\mathcal{L}_{reg}$): $L2$ distance between the Gram matrix of the orthogonally initialized weights of the model $W$ and the identity matrix $I$, multiplied by a scaling factor $\beta = 0.0001$. 
        \item DeepPrint loss ($\mathcal{L}_{ID}$): $L2$ distance between the DeepPrint embedding extracted from the source fingerprint image, $R$, and the DeepPrint embedding extracted from the reconstructed fingerprint image, $\hat{R}$. 
        \item Image/pixel loss ($\mathcal{L}_{i}$): L2 loss between the source fingerprint image, $I$, and reconstructed fingerprint image, $\hat{I}$.
        \item Overall loss ($\mathcal{L}_{gen_{e}}$) for $G_{e}(\cdot)$: $\lambda_{1}=1$, $\lambda_{2}=1$ and $\lambda_{3}=10$.
            \begin{equation}
            \label{eq:overall_loss_G_dp}
            \mathcal{L}_{gen_{e}} = \lambda_1\mathcal{L}_{A} +
            \lambda_2\mathcal{L}_{ID} + \lambda_3\mathcal{L}_{i}
            \end{equation}
        \item Overall loss ($\mathcal{L}_{disc_e}$) for $D_{A}(\cdot)$:.
            \begin{equation}
            \label{eq:overall_loss_D_dp}
            \mathcal{L}_{disc_{e}} = \mathcal{L}_{A}
            \end{equation}
    \end{enumerate}
    
    For training the DeepPrint template to image inversion model, we used the aforementioned collection of 447,988 rolled fingerprint impressions from the MSP dataset. The model was trained on a server containing two RTX 1080 TI GPUs, each containing 11 GB of memory. The generator learning rate was set at 0.0001 and the discriminator learning rate was set at 0.0001. The Generator was updated for three iterations for each iteration the Discriminator was updated.

    \begin{table}[H]
    \renewcommand{\arraystretch}{1.3}
    \caption{DeepPrint template to image inversion encoder-decoder architecture}
    \label{dp_inversion_architecture}
    \centering
    \begin{tabular}{>{\centering\arraybackslash}m{0.25\textwidth}||>{\centering\arraybackslash}m{0.15\textwidth}}
    \hline
     \textbf{Layer} &  \textbf{Output Dimensions} \\
    \hline
    \hline
    Input & $192$ \\
    \hline
    Fully Connected  & $512$ \\
    \hline
    Fully Connected  & $768$ \\
    \hline
    Reshape  & $4 \times 4 \times 48$ \\
    \hline
    Enc 0: ResBlock Up  & $8 \times 8 \times 768$ \\
    \hline
    Enc 1: ResBlock Up  & $16 \times 16 \times 384$ \\
    \hline
    Enc 2: Resblock + ResBlock Up  & $32 \times 32 \times 384$ \\
    \Xhline{4\arrayrulewidth}
    Dec 0: ResBlock + ResBlock Up  & $64 \times 64 \times 384$ \\
    \hline
    Dec 1: ResBlock + ResBlock Up  & $128 \times 128 \times 192$ \\
    \hline
    Non-Local Block $(192 \times 192)$  & $128 \times 128 \times 192$ \\
    \hline
    Dec 2: ResBlock + ResBlock Up  & $256 \times 256 \times 96$ \\
    \hline
    Dec 3: ResBlock Up  & $512 \times 512 \times 48$ \\
    \hline
    Batch Normalization, ReLU  & $512 \times 512 \times 48$ \\
    \hline
    Convolution (channels=1, kernel=$3\times3$, stride=1) & $512 \times 512 \times 1$ \\
    \hline
    Tanh & $512 \times 512 \times 1$ \\
    \Xhline{2\arrayrulewidth}
    \end{tabular}
    \end{table}

\section{Experimental Results}
    Evaluation of the template inversion networks was completed by calculating the verification accuracy for both type-I and type-II attacks on four evaluation datasets, which were not seen during training. In particular, we evaluate both the minutiae inversion attack and DeepPrint inversion attack on two rolled datasets, NISTSD4 and NISTSD14, and two slap fingerprint datasets, FVC2002 DB1\_A and FVC2002 DB2\_A. For computing the verification accuracy of the attacks, three different matchers are used: (i) Verifinger v12.3 SDK matcher by Neurotechnology~\cite{verifinger}, (ii) Innovatrics ANSI\&ISO v2.4.10 SDK and (iii) DeepPrint~\cite{engelsma2019learning}. 
    
    All previous minutiae template inversion papers evaluated their methods in a ``white-box" evaluation, where the attackers have access to the target matchers and use them in supervising the template inversion process. The inversion attack performance is then reported using the same matcher which was used in designing the template inversion algorithm. A more challenging scenario, referred to as a ``black-box" evaluation, is one in which attackers design their systems using the feedback from one system and apply their method to attacking a completely unrelated system. For our experiments, we demonstrate the success of our inversion networks in both the white-box and black-box settings. 
    
    \begin{figure*}
    \begin{center}
    \includegraphics[width=\linewidth]{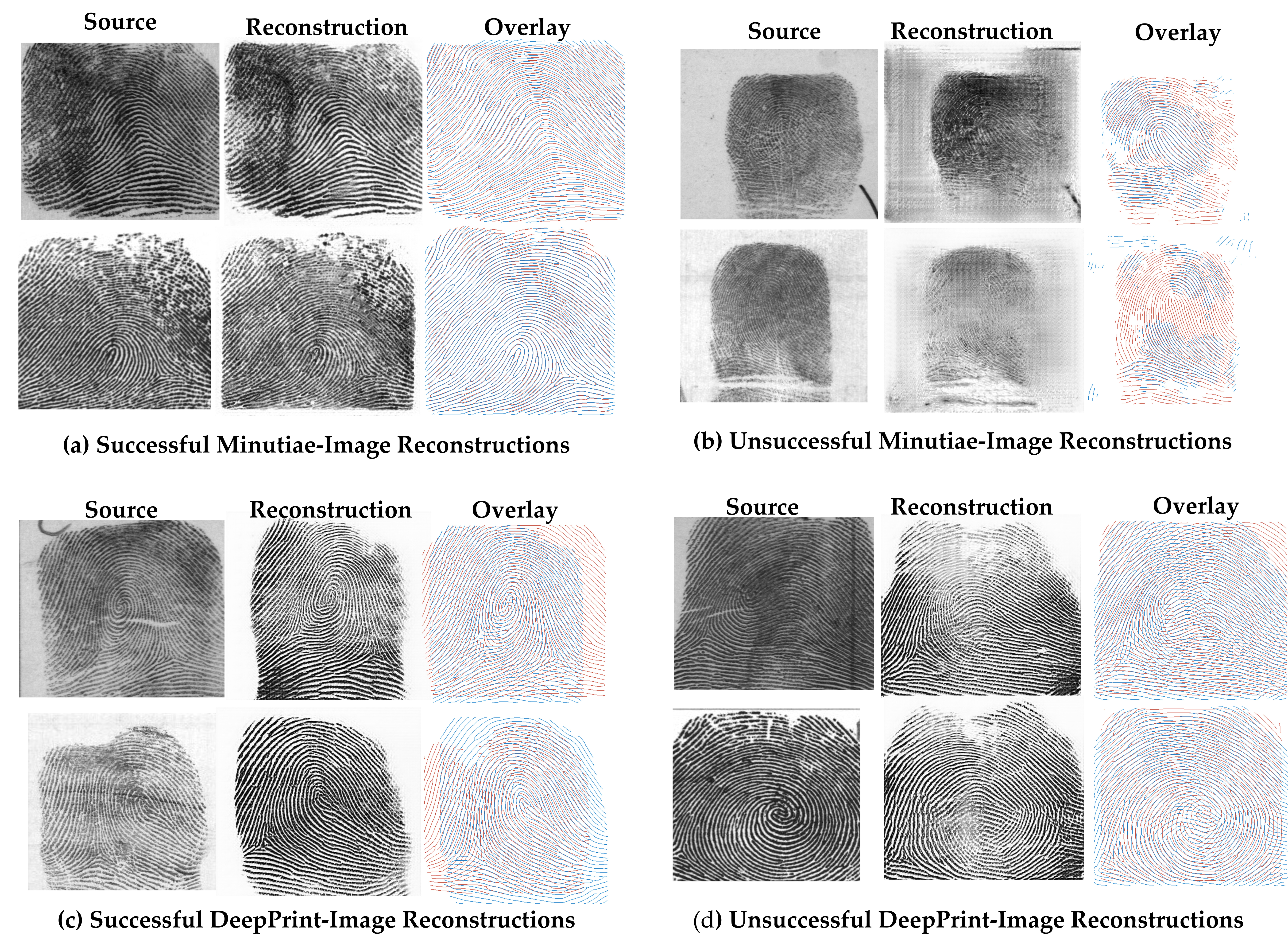} \\
    \caption{Examples of: Successful image reconstructions from minutiae templates (MSU-LatentAFIS template) (a); Unsuccessful reconstructions from minutiae (b); Successful image reconstructions from DeepPrint embeddings (c); Unsuccessful reconstructions from DeepPrint embeddings (d). The first column of each sub-figure contains the source image, the second column contains the reconstructed image, and the third column contains the ridge overlay of the reconstructed image on the source image}
    \label{fig:success_unsuccess}
    \end{center}
    \vspace{-1.5em}
    \end{figure*}
    
\subsection{Evaluation Protocol}

    To evaluate the reconstruction approach, four public domain datasets were utilized:
    \begin{enumerate}
            \item NIST SD4~\cite{sd4} is a rolled fingerprint dataset comprised of 2,000 unique fingers with two impressions per finger (a total of 4,000 fingerprint images). 
            \item NIST SD14~\cite{sd14} is a rolled fingerprint dataset comprised of 27,000 unique fingers with two impressions per finger (a total of 54,000 fingerprint images). Out of these, in the interest of time and consistency, 2,000 fingerprint pairs were selected, for a total of 4000 images. 
            \item FVC2002 DB1\_A~\cite{fvc2002} is a plain fingerprint dataset comprised of 100 unique fingers, with 8 impressions per finger (a total of 800 images).
            \item FVC2002 DB2\_A~\cite{fvc2002} is a plain fingerprint dataset comprised of 100 unique fingers, with 8 impressions per finger (a total of 800 images). 
    \end{enumerate} 
    
    Following the protocols of previous inversion attack methods, a total of 2,000 genuine match scores and 1,999,000 impostor match scores were computed for the NISTSD4 data using the available 2,000 fingerprint images. For FVC2002 DB1\_A and FVC2002 DB2\_A, genuine scores were calculated by matching each fingerprint image with its other impressions of the same finger, resulting in 2,800 genuine comparisons for each dataset and impostor scores were calculated by taking the first impression of each finger and matching it with the first impression of every other finger in the dataset, yielding 4,950 imposter match scores for both datasets. Since no previous methods evaluate on NISTSD14 and to limit the number of comparisons to a reasonable amount, just the first 2,000 fingers on NISTSD14 were included for the evaluation. This resulted in 2,000 genuine scores and 1,999,000 imposter scores.
    
    \begin{figure*}[t]
    \begin{center}
    \includegraphics[width=\linewidth]{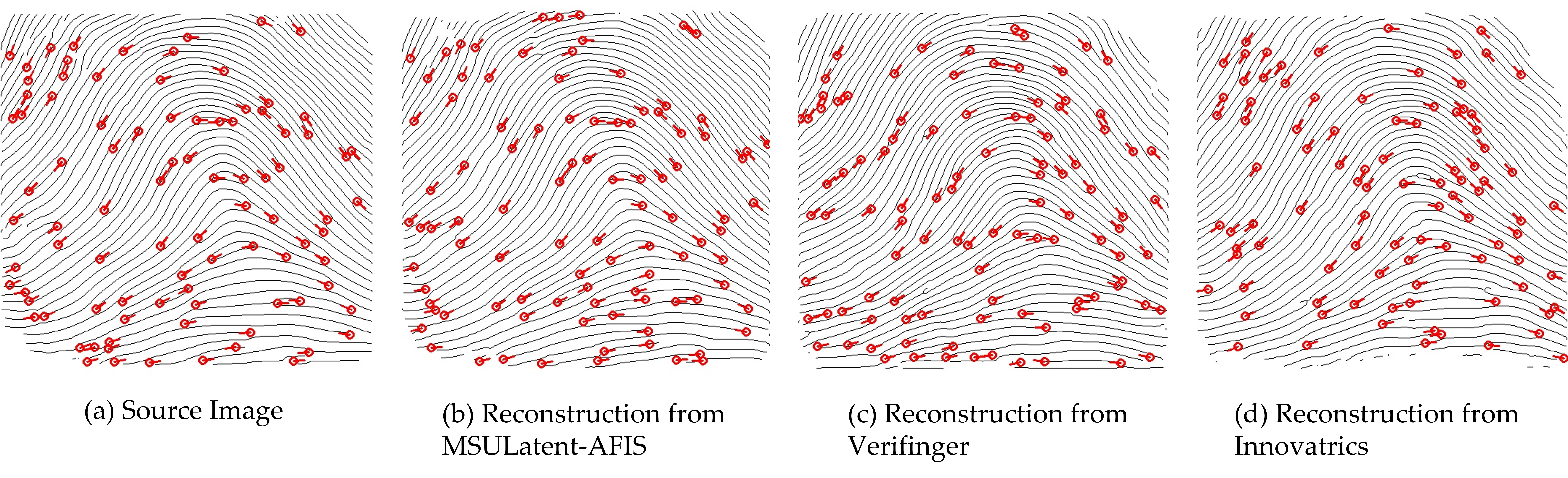} \\
    \caption{Ridge skeletons with minutiae points highlighted for: A sample fingerprint image from NIST SD4 (a); The reconstructed image of (a) via inversion of its MSU-LatentAFIS template (b); The reconstructed image of (a) via inversion of its Verifinger v12.3 minutiae template (c); The reconstructed image of (a) via inversion of its Innovatrics v2.4 minutiae template (d). The Verifinger matching scores between (b) and (a), (c) and (a) and (d) and (a) are 1199, 393 and 379 respectively, with a threshold score of 39 @ FAR of 0.01\%. }
    \label{fig:capture_fing_scores}
    \end{center}
    \vspace{-1.5em}
    \end{figure*}

\subsection{Minutiae Template Inversion Attack Performance}
    In this section, we discuss the inversion attack performance (both white-box and black-box attacks) using our minutiae-template inversion network. For a qualitative analysis, some example reconstructions are given in (a) of Figure~\ref{fig:success_unsuccess}. The corresponding source image from which the minutiae templates were derived are shown in column 1, with the reconstructed image shown in column 2. As can be observed in the ridge overlays (shown in column 3), both the level-1 (ridge pattern) and level-2 (minutiae) features of the source images are well preserved in the images reconstructed from minutiae templates. 
    
    As shown in Tables \ref{Inversion_SD4}, \ref{Inversion_SD14}, \ref{Inversion_DB1_A}, and \ref{Inversion_DB2_A}, our GAN-based minutiae template inversion network has pushed the SOTA inversion attack performance across each of the evaluation datasets in the white-box setting, in which the target matcher is used to generate the minutiae templates to be inverted. Even in the black-box setting (where the target matcher differs from the template generation SDK), the minutiae inversion attack performance is still quite high, demonstrating the capability of attackers using the templates from a compromised database from one system in attacking an unrelated fingerprint matching system. When examining Figure ~\ref{fig:capture_fing_scores}, we can see that the minutiae-based matching scores for the reconstructions from all three minutiae templates are relatively high, and that all the reconstructions captured several minutiae with a relatively high degree of accuracy. The type-I recognition accuracy was above 98\% at FAR of 0.01\% across all the white-box and black-box evaluations for minutiae-template inversion schemes matched with COTS SDKs.
    
    However, we found that the attack performance of the reconstructed images from inverting minutiae-based templates suffered to a certain degree when input to DeepPrint for matching. The drop in accuracy was particularly noticeable for type-II attacks, with the accuracy dropping below 90\% for FVC2002 DB1\_A and FVC2002 DB2\_A plain fingerprint datasets at an FAR of 0.01\%. It is important to note here that the DeepPrint model used to assess DeepPrint embedding loss during the training of the minutiae template inversion model is not the same DeepPrint model that was used to evaluate the performance of the image reconstructions from minutiae templates. These two models were trained separately to allow for a proper black-box evaluation when using the DeepPrint matcher to assess the verification accuracy. We observed that DeepPrint is more sensitive to the textural artifacts introduced in the reconstructions stage, such as missing parts of the image, blurriness, etc.; whereas the minutiae-based matchers focus on the x, y, and theta location of the minutiae. This might normally be considered to be a limitation in the recognition sense, such as local artifacts which decrease matching performance, but this actually seems to suggest that deep network-based systems are less susceptible to black-box inversion attacks from minutiae-based templates. The next section will explore whether deep learning-based recognition systems are also robust to reconstruction attacks from the deep network embeddings themselves.

\subsection{Deep-Network Template Inversion Attack Performance}

    In this section, we discuss the inversion attack performance (both white-box and black-box attacks) using our deep template inversion network. Some example reconstructions from DeepPrint templates are shown in (c) of Figure~\ref{fig:success_unsuccess}. In these examples, the level-1 features of the source fingerprint images are well preserved in the successfully reconstructed images, which is especially noticeable when observing the ridge overlay in the last column of each sub-figure. However, if we look closely at the reconstructed images, we will notice some small differences in the level-2 (e.g., minutiae) features which have deviated slightly from the source images. Due to these level-2 differences, the attack performance of the DeepPrint inverted images on each of our minutiae-based matchers is significantly lower compared to the minutiae-inversions, as can be seen in Tables \ref{Inversion_SD4}, \ref{Inversion_SD14}, \ref{Inversion_DB1_A}, and \ref{Inversion_DB2_A}. However, the DeepPrint inversion attack performance on the DeepPrint matcher itself is still quite high across each of the evaluation datasets. Thus, DeepPrint, a deep network-based fingerprint recognition system, is in fact susceptible to inversion attacks in the white-box setting. 
    
    \begin{table*}[t]
    \normalsize
    \renewcommand{\arraystretch}{1.3}
    \caption{Template inversion attack performance on \textbf{NIST SD4} using two COTS minutiae matchers and DeepPrint. Results reported as TAR (\%) @ FAR of 0.01\% for type-I (type-II) attacks.}
    \label{Inversion_SD4}
    \centering
    \begin{tabular}{||>{\centering\arraybackslash}m{0.25\linewidth}||>{\centering\arraybackslash}m{0.2\linewidth}||>{\centering\arraybackslash}m{0.2\linewidth}||>{\centering\arraybackslash}m{0.2\linewidth}||}
    \noalign{\hrule height 1.5pt}
    \backslashbox{\textbf{Template}}{\textbf{Matcher}} & \textbf{Verifinger v12.3 SDK} & \textbf{Innovatrics v2.4.10 SDK} & \textbf{DeepPrint~\cite{engelsma2019learning}} \\
    \noalign{\hrule height 1.0pt}
    \textbf{MSU-LatentAFIS~\cite{KaiMinutiae}} &                    99.90 (98.30)                 & 99.80 (97.45)                              & 99.10 (92.90)      \\
    \hline
    \textbf{Verifinger v12.3 SDK}              &                    100.0 (96.6)                 & 99.6 (90.0)                              & 35.05 (26.65)        \\
    \hline
    \textbf{Innovatrics v2.4.10 SDK} &                    99.5 (94.4)                 & 99.95 (88.65)                              & 24.95 (18.3)        \\
    \hline
    \textbf{DeepPrint~\cite{engelsma2019learning}} &                0.55 (0.75)                   & 0.35 (0.05)                                & 85.95 (68.10)     \\
    \noalign{\hrule height 1.5pt}
    \end{tabular}
    \end{table*}
    
    
    \begin{table*}[t]
    \normalsize
    \renewcommand{\arraystretch}{1.3}
    \caption{Template inversion attack performance on \textbf{NIST SD14} using two COTS minutiae matchers and DeepPrint. Results reported as TAR (\%) @ FAR of 0.01\% for type-I (type-II) attacks.}
    \label{Inversion_SD14}
    \centering
    \begin{tabular}{||>{\centering\arraybackslash}m{0.25\linewidth}||>{\centering\arraybackslash}m{0.2\linewidth}||>{\centering\arraybackslash}m{0.2\linewidth}||>{\centering\arraybackslash}m{0.2\linewidth}||}
    \noalign{\hrule height 1.5pt}
    \backslashbox{\textbf{Template}}{\textbf{Matcher}} & \textbf{Verifinger v12.3 SDK} & \textbf{Innovatrics v2.4.10 SDK} & \textbf{DeepPrint~\cite{engelsma2019learning}} \\
    \noalign{\hrule height 1.0pt}
    \textbf{MSU-LatentAFIS~\cite{KaiMinutiae}} & 99.85 (98.15)  & 99.70 (96.10) & 99.30 (94.10) \\
    \hline
    \textbf{Verifinger v12.3 SDK}              & 100.0 ( 96.85) & 99.55 (85.85) & 56.05 (43.45)   \\
    \hline
    \textbf{Innovatrics v2.4.10 SDK} & 99.65 (93.35)  & 99.9 (84.35) & 39.25 (27.0)   \\
    \hline
    \textbf{DeepPrint~\cite{engelsma2019learning}} & 0.75 (0.45)    & 0.30 (0.30)   & 96.20 (82.95)\\
    \noalign{\hrule height 1.5pt}
    \end{tabular}
    \end{table*}
    
    \begin{table*}[t]
    \normalsize
    \renewcommand{\arraystretch}{1.3}
    \caption{Template inversion attack performance on \textbf{FVC2002 DB1\_A} using two COTS minutiae matchers and DeepPrint. Results reported as TAR (\%) @ FAR of 0.01\% for type-I (type-II) attacks.}
    \label{Inversion_DB1_A}
    \centering
    \begin{tabular}{||>{\centering\arraybackslash}m{0.25\linewidth}||>{\centering\arraybackslash}m{0.2\linewidth}||>{\centering\arraybackslash}m{0.2\linewidth}||>{\centering\arraybackslash}m{0.2\linewidth}||}
    \noalign{\hrule height 1.5pt}
    \backslashbox{\textbf{Template}}{\textbf{Matcher}} & \textbf{Verifinger v12.3 SDK} & \textbf{Innovatrics v2.4.10 SDK} & \textbf{DeepPrint~\cite{engelsma2019learning}} \\
    \noalign{\hrule height 1.0pt}
    \textbf{MSU-LatentAFIS~\cite{KaiMinutiae}} & 100.0 (99.21) & 100.0 (98.36) & 77.00 (46.79) \\
    \hline
    \textbf{Verifinger v12.3 SDK}              & 99.88 (97.43) & 99.5 (90.39) & 12.63 (6.93)   \\
    \hline
    \textbf{Innovatrics v2.4.10 SDK} & 99.38 (93.11) & 98.75 (84.82) & 5.38 (2.55)   \\
    \hline
    \textbf{DeepPrint~\cite{engelsma2019learning}}  & 0.50 (0.27)   & 0.63 (0.43)   & 31.62 (13.70)\\
    \noalign{\hrule height 1.5pt}
    \end{tabular}
    \end{table*}
    
    \begin{table*}[t]
    \normalsize
    \renewcommand{\arraystretch}{1.3}
    \caption{Template inversion attack performance on \textbf{FVC2002 DB2\_A} using two COTS minutiae matchers and DeepPrint. Results reported as TAR (\%) @ FAR of 0.01\% for type-I (type-II) attacks.}
    \label{Inversion_DB2_A}
    \centering
    \begin{tabular}{||>{\centering\arraybackslash}m{0.25\linewidth}||>{\centering\arraybackslash}m{0.2\linewidth}||>{\centering\arraybackslash}m{0.2\linewidth}||>{\centering\arraybackslash}m{0.2\linewidth}||}
    \noalign{\hrule height 1.5pt}
    \backslashbox{\textbf{Template}}{\textbf{Matcher}} & \textbf{Verifinger v12.3 SDK} & \textbf{Innovatrics v2.4.10 SDK} & \textbf{DeepPrint~\cite{engelsma2019learning}} \\
    \noalign{\hrule height 1.0pt}
    \textbf{MSU-LatentAFIS~\cite{KaiMinutiae}}                        & 99.75 (98.91) & 99.38 (97.57) & 95.63 (74.0) \\
    \hline
    \textbf{Verifinger v12.3 SDK}              & 100.0 (96.0) & 98.38 (85.41) & 29.13 (17.57)   \\
    \hline
    \textbf{Innovatrics v2.4.10 SDK} & 99.0 (91.57) & 99.25 (78.8) & 25.25 (8.27)  \\
    \hline
    \textbf{DeepPrint~\cite{engelsma2019learning}}  & 0.13 (0.018)   & 0.00 (0.036)   & 82.63 (50.89)\\
    \noalign{\hrule height 1.5pt}
    \end{tabular}
    \end{table*}

\section{Discussion}
    Despite the success of the inversion attacks, there are still instances where it fails. Some concrete examples of failed type-II attacks on minutiae inversions and DeepPrint inversions are shown in (b) and (d) of Figure~\ref{fig:success_unsuccess}, respectively. In most instances, the failures can be attributed to one of the following factors:
    
    \begin{enumerate}
        \item Lack of sharpness/poor image quality of the source image
        \item Blurry patches taking up a majority of the reconstructed image or concentrating near the Level I features in the reconstructed image
        \item Presence of spurious minutiae in the reconstructed image.
    \end{enumerate}
    
    
    
    Clearly, the inversion attack performance is best when inverting the minutiae templates from MSU's Latent Minutiae Extractor, compared to inverting the minutiae from Verifinger or Innovatrics. The drop in performance is likely due to our use of MSU's Latent Matcher for both obtaining the ground truth for our minutiae inversion network as well as providing supervision to the network via the minutiae map loss during training. For a better understanding of the difference in performance, Figure~\ref{fig:capture_fing_scores} shows some example reconstructions from each of the three minutiae extractors. We can see that when inverting from minutiae provided by MSU's Latent Minutiae Extractor, the images are smooth and have relatively few artifacts; whereas inverting from either Verifinger or Innovatrics minutiae, there seem to be quite a few broken ridges and places where the ridge lines are missing. The difference between the minutiae distribution output by MSU's extractor, which the inversion network was trained with, and both Verifinger and Innovatrics seems to be the source of the issue. In some cases, there are minutiae which are detected by MSU's extractor which are missing from either Verifinger or Innovatrics and vice versa, which degrades the inversion performance.
    
    One of the main questions which this study aimed to address is whether deep network-based embeddings suffer from the same invertibility threat that minutiae-based templates are prone to. The results seem to indicate that deep network-based embeddings, such as DeepPrint, do suffer from inversion attacks when the target matcher is the same as the system used to create the inverted template. However, we found that the reconstructed images from DeepPrint templates were much less successful in attacking minutiae-based matchers. For instance, while a type-I attack performed on the inversion of a Verifinger minutiae template on NIST SD4 yielded a 100.0\% TAR at FAR of 0.01\% accuracy using the Verifinger 12.3 SDK matcher, the same attack performed for deep template inversion only yielded a type-I attack accuracy of 0.55\% TAR at FAR of 0.01\%. However, the DeepPrint inversion attack on the DeepPrint matcher itself still manages to yield a 85.95\% type-I and a 68.10\% type-II attack performance.
    
    \begin{figure}
    \centering
    \includegraphics[width=\linewidth]{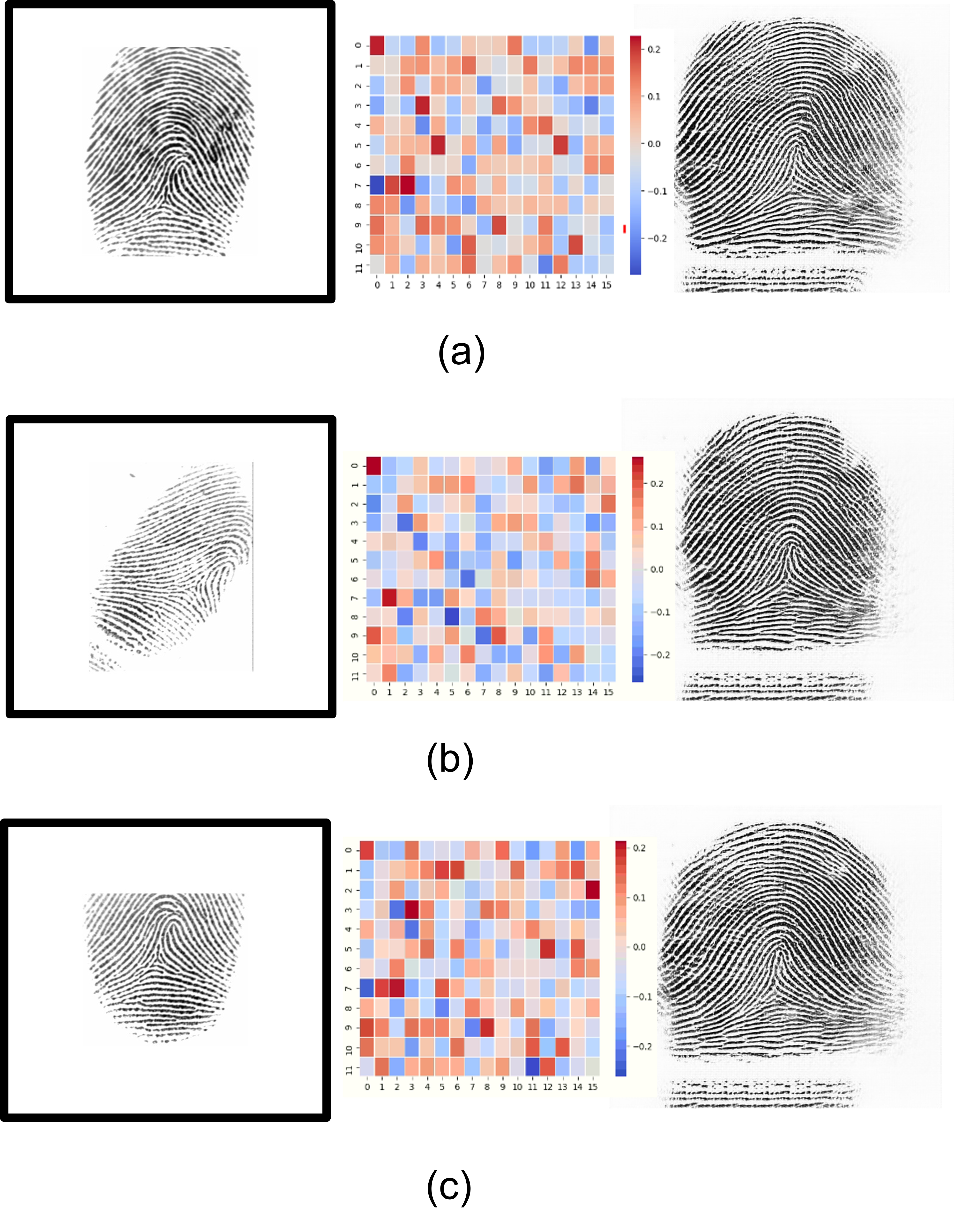} \\
    \caption{Comparison of deep embedding inversions of three different impressions (a), (b) and (c) of the same fingerprint image from FVC2002 DB1\_A. In each row, the image on the left represents the source fingerprint, the heatmap in the middle represents the DeepPrint embedding of the source fingerprint, and the image on the right represents the corresponding reconstruction from the source DeepPrint embedding}
    \label{fig:dp_comp}
    \vspace{-1.5em}
    \end{figure}
    
    We found that even though the level-1 features of the fingerprints can be easily reconstructed from DeepPrint templates as observed in Figure~\ref{fig:dp_comp}, the level-2 features (e.g., minutiae) are not reliably preserved in the reconstructed image. This subsequently leads to unsuccessful attacks against minutiae-based systems. The difficulty in reconstructing the level-2 features stems from the fact that DeepPrint is trained to be invariant to affine transformations (via random translation and rotation augmentations) causing it to ignore the original spatial location of the source fingerprint image. The lack of spatial correspondence between the source and reconstructed images makes it difficult to apply further supervision to preserve the minutiae in the reconstructed images. Indeed, looking at Figure~\ref{fig:dp_comp}, we can see that several impressions of the same finger, which have varying translation and rotation, all produce virtually the same DeepPrint embedding and identical reconstructions located at the same spatial location within the image. It may also be the case that DeepPrint is not encoding complete level-2 information (e.g., minutiae locations) as precisely as a minutiae representation itself. We leave it to future work whether we can achieve spatial correspondence between the ground truth source image and reconstructed images either through a pre-alignment of the ground truth images or by incorporating a Spatial Transformer Network to learn the alignment during training of the inversion network. Such an alignment may help to better supervise the reconstruction of level-2 features.

\section{Conclusions and Future Work}
    In this work, we demonstrated the feasibility of inverting deep template fingerprint embeddings. We found that deep template embeddings are indeed susceptible to inversion attacks; albeit, only in the white-box setting. In particular, reconstructed images from deep template embeddings could reliably match with corresponding images by the same deep network used to create the embeddings; however, could not match reliably with the source images via the minutiae-based matchers used in this study. In comparison, reconstructed images from minutiae-based templates could reliably match with the source images for both white-box and black-box attacks. Furthermore, we observed that black-box attacks on our deep network-based matcher, DeepPrint, from Verifinger or Innovatrics minutiae-based templates were much less successful compared to the attacks on the minutiae-based matchers. These results suggest that deep network fingerprint embeddings may be more secure due to their robustness against black-box attacks from templates derived from an unrelated, database of compromised fingerprint templates. Future work would investigate whether it is possible to improve the Level-2 details in the fingerprint images reconstructed from deep template embeddings to improve the inversion attack performance across minutiae-based matchers.

\section{Acknowledgment}
    The authors would like to thank Dr. Joshua J. Engelsma for his valuable discussions and contributions to this work.

{\small
\bibliographystyle{ieeetr}
\bibliography{citations}
}

\begin{IEEEbiography}[{\includegraphics[width=1in,height=1.25in,clip,keepaspectratio]{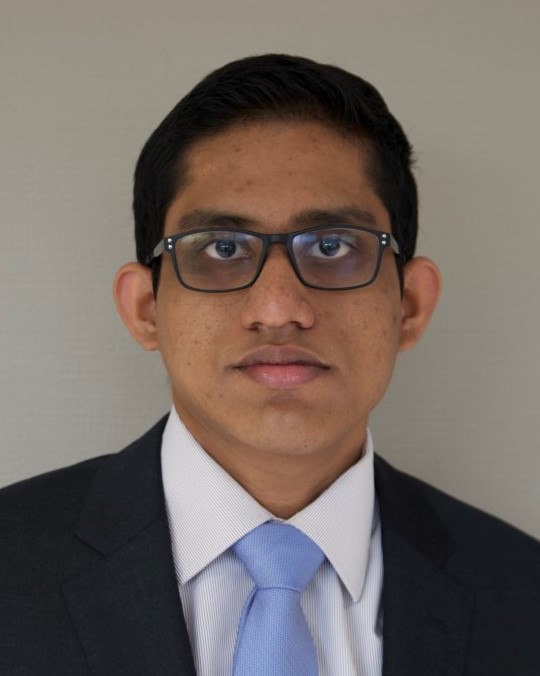}}]{Kanishka P. Wijewardena}
received his B.S. degree in Computer Engineering from Michigan State University, East Lansing, Michigan, in 2021. He is currently pursuing his Ph.D. degree in the Department of Computer Science and Engineering at Michigan State University. His primary research interests are in the areas of biometrics, machine learning and computer vision.
\end{IEEEbiography}

\begin{IEEEbiography}[{\includegraphics[width=1in,height=1.25in,clip,keepaspectratio]{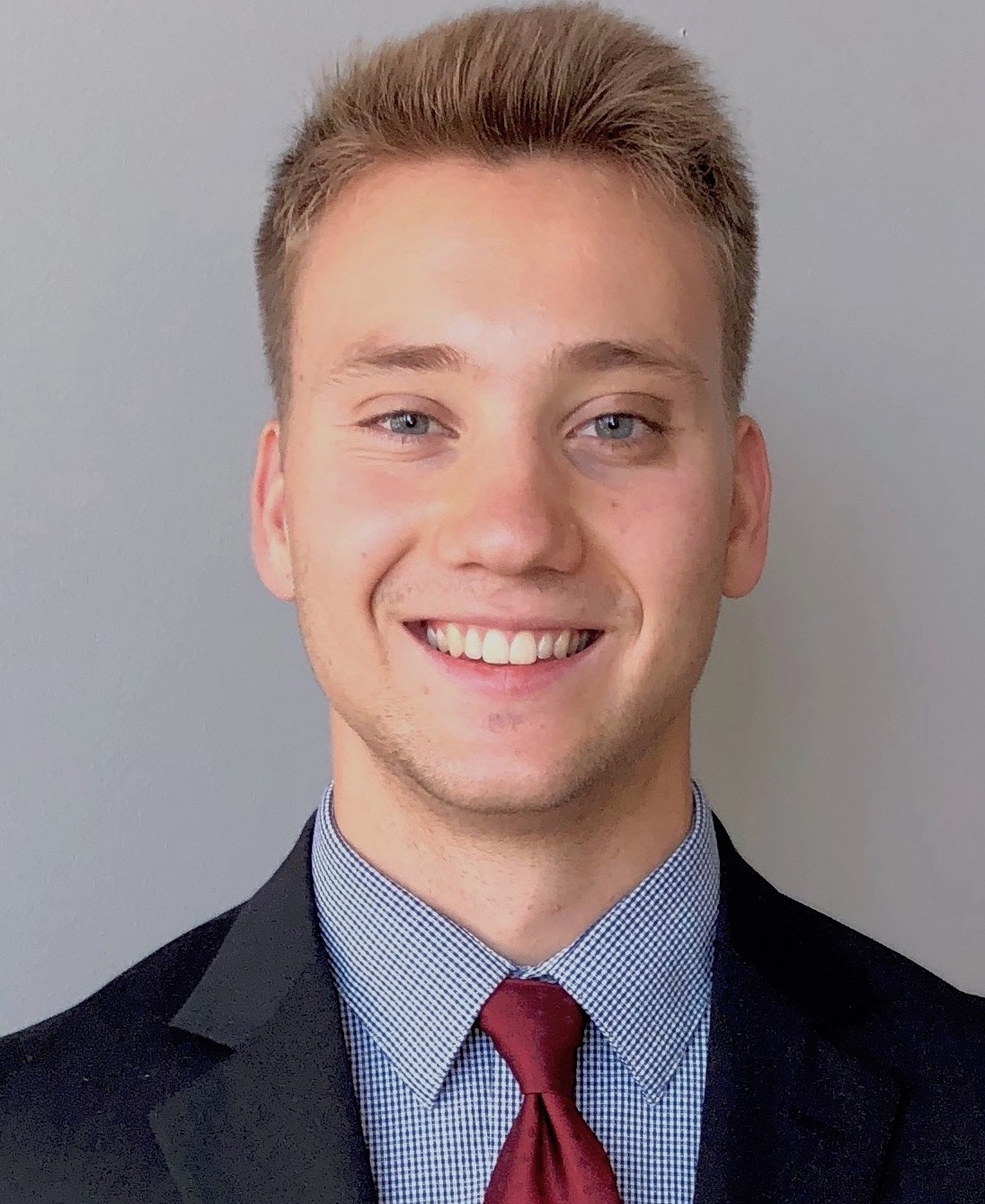}}]{Steven A. Grosz}
received his B.S. degree in Electrical Engineering from Michigan State University, East Lansing, Michigan, in 2019. He is currently pursuing his Ph.D. degree in the Department of Computer Science and Engineering at Michigan State University. His primary research interests are in the areas of machine learning and computer vision with applications in biometrics.
\end{IEEEbiography}

\begin{IEEEbiography}[{\includegraphics[width=1in,height=1.25in,clip,keepaspectratio]{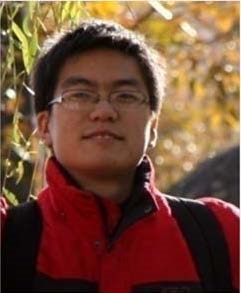}}]{Kai Cao}
received the Ph.D. degree from the Key Laboratory of Complex Systems and Intelligence Science, Institute of Automation, Chinese Academy of Sciences, Beijing, China, in 2010. He was a Post Doctoral Fellow in the Department of Computer Science \& Engineering, Michigan State University. He is now a Senior Research Scientist at Goodix in San Diego, CA. His research interests include biometric recognition, computer vision, image processing and machine learning.
\end{IEEEbiography}

\begin{IEEEbiography}[{\includegraphics[width=1in,height=1.25in,clip,keepaspectratio]{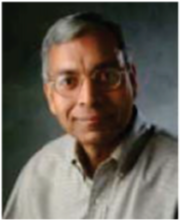}}]{Anil K. Jain}
is a University distinguished professor in the Department of Computer Science and Engineering at Michigan State University. His research interests include pattern recognition and biometric recognition. He served as the editor-in-chief of the IEEE Transactions on Pattern Analysis and Machine Intelligence. He is a member of the United States National Academy of Engineering and foreign member of the Indian National Academy of Engineering, and Chinese Academy of Sciences.
\end{IEEEbiography}

\end{document}